\documentclass[sigconf,review=False]{acmart}
\AtBeginDocument{%
  \providecommand\BibTeX{{%
    \normalfont B\kern-0.5em{\scshape i\kern-0.25em b}\kern-0.8em\TeX}}}

\setcopyright{acmcopyright}
\copyrightyear{2021}
\acmYear{2021}
\acmDOI{10.1145/1122445.1122456}

\acmConference[Conference'21]{In Proceedings of ACM Conference}{ 2021.ACM}{ACM, NY}
\acmPrice{15.00}
\acmISBN{978-1-4503-XXXX-X/21/06}

\usepackage[noend]{algpseudocode}
\begin{document}

\title{Video 3D Sampling for Self-supervised Representation Learning}


\author{Wei Li, Dezhao Luo, Bo Fang, Yu Zhou, Weiping Wang}
\email{liwei1@iie.ac.cn, luodezhao@iie.ac.cn, fangbo@iie.ac.cn, zhouyu@iie.ac.cn, wangweiping@iie.ac.cn}
\affiliation{
  \institution{Institute of Information Engineering, Chinese Academy of Sciences}
  \streetaddress{}
  \city{}
  \state{}
  \country{}
  \postcode{}
}

\begin{abstract}
   Most of the existing video self-supervised methods mainly leverage temporal signals of videos, ignoring that the semantics of moving objects and environmental information are all critical for video-related tasks. In this paper, we propose a novel self-supervised method for video representation learning, referred to as Video 3D Sampling (V3S). In order to sufficiently utilize the information (spatial and temporal) provided in videos, we pre-process a video from three dimensions (width, height, time). As a result, we can leverage the spatial information (the size of objects), temporal information (the direction and  magnitude of motions) as our learning target. In our implementation, we combine the sampling of the three dimensions and propose the scale and projection transformations in space and time respectively. The experimental results show that, when applied to action recognition, video retrieval and action similarity labeling, our approach improves the state-of-the-arts with significant margins. 
\end{abstract}




\maketitle

\begin{figure}[t]
\centering
\includegraphics[width=0.9\columnwidth]{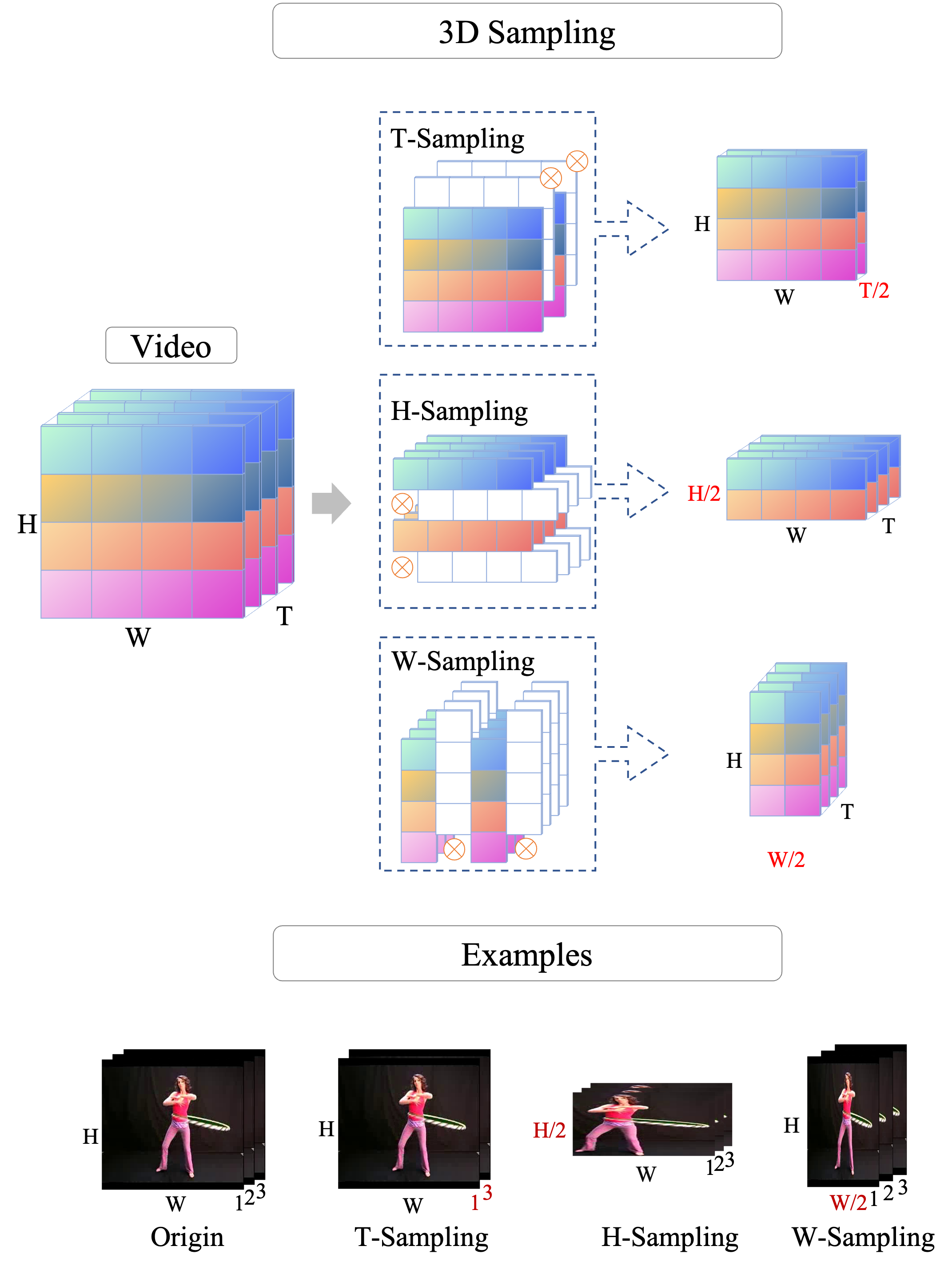}
\caption{V3S novelly extracts video information from three dimensions (width, height, time) as supervised labels. For T-Sampling, frames of the video are skipped with intervals. For H-Sampling, pixels in the frame are removed in rows, which stretches the image along its width. Accordingly, W-Sampling stretches the image along its height.
}
\label{fig:introduction}
\end{figure}


\section{Introduction}
With the development of convolutional neural networks (CNNs), many computer vision tasks have achieved great progress in recent years. Even though supervised learning shows promising results, training CNNs needs large labeled datasets like ImageNet \cite{deng2009imagenet} and Kinetics \cite{kay2017kinetics}.  
However, it is very expensive and time-consuming to annotate large-scale datasets, especially for complex annotation tasks. To explain it further, the semantic segmentation task requires the category of each pixel, and action detection requires the borders and category of
each action instance. From this point of view, training CNNs in a self-supervised manner is of great significance.

Recently, self-supervised learning is proposed to utilize unlabeled data to learn representations. Typically, self-supervised methods automatically generate labels from raw data itself and design a proxy task to predict the labels. In this manner, CNNs are encouraged to learn representative features without manual annotations, and the learned features can be used to finetune downstream tasks.  

For image self-supervised learning, early methods intended to learn representations by predicting the angle of images \cite{gidaris2018unsupervised}, the relative patch location \cite{noroozi2016unsupervised}, or the removed region of the image \cite{pathak2016context}.
For video self-supervised learning, a growing number of researches focused on modeling temporal transformations of videos. {\cite{lee2017unsupervised,misra2016shuffle,xu2019self}}  shuffled video frames or clips and utilized the original order as the learning target. SpeedNet \cite{benaim2020speednet}, PRP \cite{yao2020video}, PacePred \cite{wang2020self} randomly sped up the video and predicted the speed of the video to learn representations. {\cite{jenni2020video}} investigated 4 temporal transformations ({\it e.g. speed, random, periodic, and warp}) and demonstrated their effectiveness to guide representation learning. 

While promising results have been achieved, the above works still have some drawbacks. First of all, they do not sufficiently exploit the information provided by videos. The motion of an object contains two parameters:  magnitude and direction. For temporal representation learning, speed/playback rate based methods \cite{benaim2020speednet,yao2020video} modify the magnitude of motion and use it as the learning target. They ignore the moving direction of objects. Moreover, previous methods in video self-supervised learning tend to focus on temporal feature learning. Most of them do not exploit tasks for spatial representation learning. However, appearance information, including the semantics of the moving objects and environmental information, are also essential for video-related tasks. 
Secondly, the preprocessing strategies proposed by previous methods tend to destroy the video's semantic structure, resulting in unreasonable content. For example, order-based method \cite{xu2019self} disrupts the motion pattern and uses the original order as the learning target. However, shuffling video frames will seriously affect the content semantics. While (cubic puzzle \cite{kim2019self}, VCP \cite{luo2020video}) design spatial labels, they severely destroy the spatial structure. Therefore, we intend to investigate how motion direction, as well as spatial semantics, contribute to video representation learning in a simple-yet-effective manner. 

In this paper, we propose a novel self-supervised representation learning approach referred to as video 3D Sampling (V3S). As shown in Fig. \ref{fig:introduction}, our goal is to make full use of the spatio-temporal information in videos without changing its semantics. To learn temporal features, we leverage the direction and  magnitude  of motions as the learning target. To learn spatial features, we apply spatial scale and spatial projection to modify the size of objects and the direction of motions. Accordingly, we apply temporal scale and temporal projection for temporal representation learning.  To avoid losing too many frames in the speed-up process, which might lead to the loss of the original semantic information, we propose a progressive fast-forward sampling strategy. In this way, we transform the task of predicting the speed to that of predicting the changing pattern of the speed. As a self-supervised method, the transformations mentioned above are used as supervisory signals. In order to fully integrate spatial and temporal features, we set V3S as a multi-task learning framework.

The main contributions of this work are summarized as follows:

\begin{itemize}
    \item We propose video 3D sampling to learn spatio-temporal representations. Targeting at exploiting the information a video contains comprehensively, V3S samples video information at all three dimensions of width, height, and time. 
    
    \item For spatial representation learning, we modify the aspect ratio of objects as the learning target. For temporal representation learning, we exploit the direction of motions and propose to speed up videos progressively.

    \item We verify V3S's effectiveness on 4 backbones (C3D, R3D, R(2+1)D, S3D-G) and 3 tasks (action recognition, video retrieval, action similarity labeling), which demonstrates that V3S improves the state-of-the-arts with significant margins.
\end{itemize}

\section{Related Work}
In this section, we first introduce video representation learning in self-supervised manners. Then we introduce the recent development of video action recognition.

\subsection{Self-Supervised Representation Learning}
By generating pseudo labels from the raw data, self-supervised learning methods can learn rich representations without leveraging expensive human-annotated labels. 
Self-supervised image representation learning has witnessed rapid progress. Various proxy tasks have been proposed such as jigsaw puzzle \cite{noroozi2016unsupervised}, rotation prediction \cite{gidaris2018unsupervised}, colorization \cite{zhang2016colorful}, inpainting \cite{pathak2016context}, and context prediction \cite{doersch2015unsupervised}, to name a few. Video representation learning can be categorized into temporal representation learning and spatiao-temporal representation learning.

\subsubsection{Temporal Representation Learning}
Existing self-supervised learning methods in image classification can be directly applied  to video representation learning due to the fact that video frames are images in essence. Moreover, distinct temporal information of videos was demonstrated effective for many vision tasks (e.g. action recognition). Prior works have explored the temporal ordering of the video frames as a supervisory signal.

Based on 2D-CNNs, \cite{lee2017unsupervised,misra2016shuffle} took temporally shuffled frames as inputs and trained a ConvNet to sort the shuffled sequences. \cite{wei2018learning} exploited the arrow of time as a supervisory signal. In \cite{fernando2017self}, an odd-one-out network was proposed to identify the unrelated or odd clips from a set of otherwise related clips. Recently, video self-supervised learning performance has been largely boosted due to 3D-CNNs (e.g. C3D \cite{tran2015learning}, S3D-G \cite{xie2018rethinking}). 

VCOP \cite{xu2019self} extended the 2D frame ordering pretext tasks to 3D clip ordering. In SpeedNet \cite{benaim2020speednet} and PRP \cite{yao2020video}, the network was trained to predict the video playback rate, which was proved to be effective in learning the foreground moving objects. Similarly, \cite{wang2020self} introduced a pace prediction task that fused the novel option of slow motion. Furthermore, \cite{jenni2020video} investigated multiple different temporal transformations (\emph{speed}, \emph{periodic}, \emph{warp}, \emph{etc.}) to build a useful representation of videos for action recognition.

\begin{figure*}
     \centering
     \includegraphics[width=2.1\columnwidth]{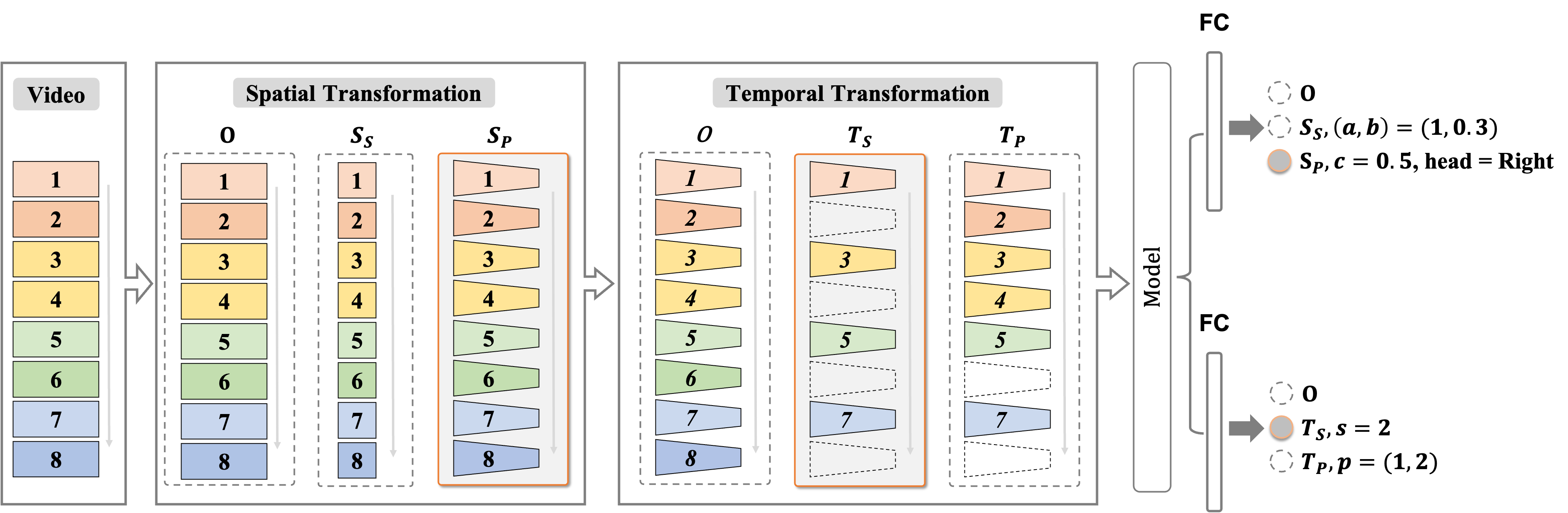}
     \caption{Illustration of V3S framework. Given a raw video, spatial and temporal transformations are carried out on it in turn. A model is applied to extract the feature of the transformed video, then the feature is fed to two FC layers to predict the specific category in space and time separately. $O$ denotes the original video, $S_S$ denotes the spatial scale transformation, $S_P$ spatial projection, $T_S$ temporal scale, $T_P$ temporal projection. The details of the parameters are discussed in Sec. \ref{sec:method}.}
     \label{fig:method}
 \end{figure*}

\subsubsection{Spatio-Temporal Representation Learning}
Despite the success in temporal representation learning, spatial transformations applied to video frames are still necessary. It can be interpreted that spatial representation learning focuses more on the appearance of objects while temporal representation learning tends to learn motion patterns. Several spatio-temporal self-supervised methods have been proposed recently. \cite{kim2019self} trained 3D-CNNs by completing space-time cubic puzzles. \cite{jing2018self} proposed 3DRotNet using rotation angel as a supervisory signal, which extended the rotation operation from images to videos. \cite{wang2019self} proposed to regress both motion and appearance statistics along spatial and temporal dimensions for representation learning. VCP \cite{luo2020video} designed the video cloze procedure task to learn the spatio-temporal information of videos to full advantage. In addition, future frame prediction \cite{han2019video,han2020memory}, was usually a considered approach for video representation learning.


Self-supervised learning combined with contrastive learning currently has demonstrated promising results, e.g. \cite{wang2020self}. In this paper, we focus on designing pure proxy tasks and leave the potential extension to contrastive learning as future research.

\subsection{Video Action Recognition}

Action recognition is one of the most important tasks for video understanding. It takes a video clip as the input and outputs the specific action category of the video. Since the dynamic information is complex to understand, action recognition is challenging. 

Based on the 2D CNNs feature extractor, \cite{simonyan2014two} proposed two-stream convolutional networks where the results of RGB stream and optical flow stream were fused. TSN \cite{wang2016temporal} extracted multiple clips for a video and utilized the whole action video level supervision. \cite{zhou2018temporal} built temporal dependencies among video frames for action recognition.

Recently, 3D CNNs feature extractors have attracted much attention due to their strong ability for temporal modeling.  C3D \cite{tran2015learning} designed 3D convolutional kernels, which can model spatial and temporal features simultaneously. R3D \cite{tran2018closer} extended C3D with ResNet\cite{he2016deep}. S3D-G \cite{xie2018rethinking}  replaced 3D CNNs at the bottom of the network with low-cost 2D convolutions.

\section{Methods}
\label{sec:method}
 Recent methods use sampling interval  \cite{benaim2020speednet} or clip order \cite{xu2019self} as the learning target to learn temporal features.  Specifically, \cite{benaim2020speednet} generates speed-up videos by interval sampling, which enhances the  magnitude   of the motions. 
 In this work, in addition to learning the  magnitude of motions, we also leverage the direction of motions as one of our learning targets. Moreover, spatial semantics are used in our methods, which are ignored by previous methods.

Our goal is to encourage CNNs to learn rich spatial and temporal representations. For spatial representation learning, we apply scale and projection transformations. For temporal representation learning, we further extend scale and projection transformation on the temporal dimension. We are going to describe these transformations in the following.
\subsection{Spatial Transformation}
\label{transformation}
To encourage the model to learn spatial representations, we design two transformations on the appearance of video clips. 
Let $I(x,y)$ be the original frame, $I(u,v)$ be the transformed frame. There is a conversion formula  which maps ($x$,$y$) to ($u$,$v$):
 $$   
u =\frac{m_{0}*x+m_{1}*y+m_{2}}{m_{6}*x+m_{7}*y+1},
v =\frac{m_{3}*x+m_{4}*y+m_{5}}{m_{6}*x+m_{7}*y+1}.
$$
$$
\begin{bmatrix}
m_{0}& m_{1}& m_{2}
\\
m_{3}& m_{4}& m_{5}
\\
m_{6}& m_{7}& 1
\end{bmatrix}
is calculated by solving linear system:
$$
$$
\begin{bmatrix}
x_{0}& y_{0}& 1 & 0 & 0 & 0 & -x_{0}*u_{0} & -y_{0} * u_{0}
\\
x_{1}& y_{1}& 1 & 0 & 0 & 0 & -x_{1}*u_{1} & -y_{1} * u_{1}\\
x_{2}& y_{2}& 1 & 0 & 0 & 0 & -x_{2}*u_{2} & -y_{2} * u_{2}\\
x_{3}& y_{3}& 1 & 0 & 0 & 0 & -x_{3}*u_{3} & -y_{3} * u_{3}\\
0 & 0 & 0 &x_{0}& y_{0}& 1 &  -x_{0}*v_{0} & -y_{0} * v_{0}\\
0 & 0 & 0 &x_{1}& y_{1}& 1 &  -x_{1}*v_{1} & -y_{1} * v_{1}\\
0 & 0 & 0 &x_{2}& y_{2}& 1 &  -x_{2}*v_{2} & -y_{2} * v_{2}\\
0 & 0 & 0 &x_{3}& y_{3}& 1 &  -x_{3}*v_{3} & -y_{3} * v_{3}
\\
\end{bmatrix} 
\cdot
\begin{bmatrix}
m_0\\
m_1\\
m_2\\
m_3\\
m_4\\
m_5\\
m_6\\
m_7\\
\end{bmatrix}
=
\begin{bmatrix}
u_0\\
u_1\\
u_2\\
u_3\\
v_0\\
v_1\\
v_2\\
v_3\\
\end{bmatrix}
$$
, where $(x_i,y_i),(u_j,v_j)(i,j=1,2,3,4)$ are the coordinates of the four vertices of the original and the transformed image. To explain it in detail,  we firstly obtain the coordinates $(x,y)$ of four vertices in the original frame, and then calculate the coordinates  $(u,v)$ of vertices after transformation. Then, $m_i (i=0,...,7)$ is calculated by the above formula. Thus, every pixel in the original image can find the corresponding location in the transformed frame.

To make it simple, we set $(x_0,y_0) = (0,0), (x_1,y_1) = (0,H), (x_2,y_2)$ $= (W,H), (x_3,y_3) = (W,0)$, where $W,H$ denotes the width and the height of the original image. We also set $(u_0,v_0) = (0,0)$. In the following, we are going to describe the detail of the transformations.



\textbf{Scale:} 
In order to change the size (aspect ratio) of the object, we modify the height or width of the video frame. It should be noted that we do not scale the height and width equally (with the same rate), because this only changes the resolution of the image, which is trivial to learn. Moreover, if we change the aspect ratio of the object, and the network can still learn its original proportion, it demonstrates that the network has learned the semantic information of the object.

In scale, we set $(u_1,v_1) = (0,b*H), (u_2,v_2) = (a*W,b*H), (u_3,v_3) = (a*W,0)$. It denotes that the width and height of the transformed video frame is $a$ and $b$ times of that of the original video frame. In our implementation, $(a,b)$ is the hyperparameter and the learning target.  Fig: \ref{fig:method}
shows an example of $(a,b) = (1,0.3) $.

\textbf{Projection}: Projection transforms a cuboid into a trapezoid. The head end of the trapezoid is smaller than the tail end, which has the effect of expanding objects close to the camera and shrinking distant objects. In order to change the sizes of objects in different regions with different rates, we apply projection on the video frame.

To transform the frame to a trapezoid, we randomly choose one side as the head end and shorten the length. For example, we take the right side as the head end and set $(u_1,v_1) = (0,H)$, $(u_2,v_2)$ =$ (W,(H+c*H)/2),(u_3,v_3) = (W,(H-c*H)/2)$.
It denotes that the transformed frame is a trapezoid which takes the left side as the bottom end and the right as the head end. The length of the head end is =$c*H$.  For spatial projection, $c$ and the head end side is the learning target. Fig. \ref{fig:method} shows an example of $c = 0.5$ and the head end is the right side.

\begin{figure}
     \centering
     \includegraphics[width=1.0\columnwidth]{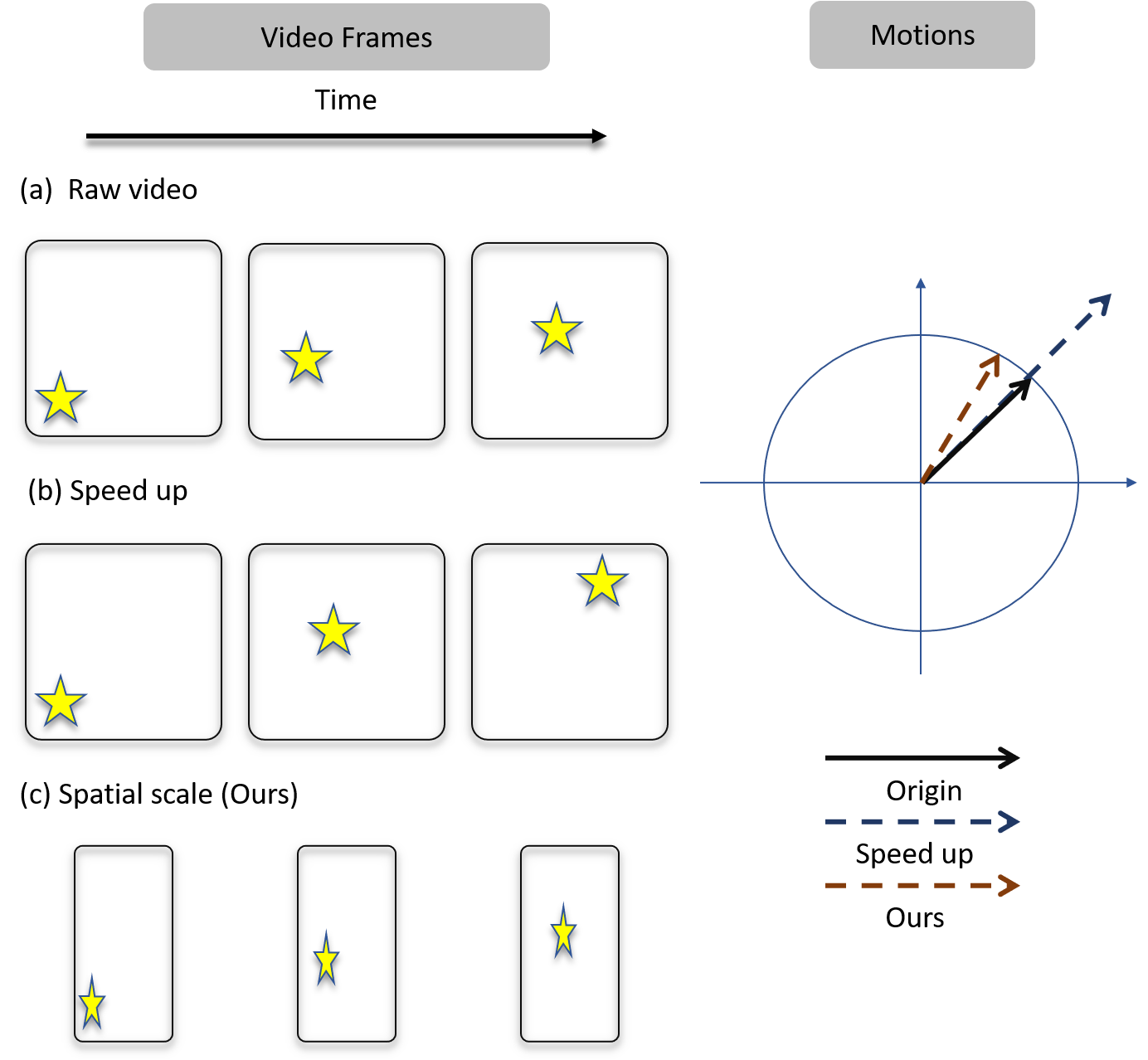}
     \caption{The left (video frames) shows moving objects in videos. The right (motions) shows the magnitude and direction of the motions. V3S can leverage the direction of motions as the learning target. }
     \label{fig:direction}
 \end{figure}

Through spatial transformations, we can change the size (aspect ratio) of the objects uniformly (scale) or non-uniformly (projection) whilst maintaining the semantic information. 
 Fig. \ref{fig:direction} shows the example of a three-frame video with an object moving in a straight line.  Through spatial transformations, we can modify the direction of motions. 

\subsection{Temporal Transformation}
To encourage the model to learn rich temporal representations, two temporal transformations on video clips are carried out. We design different frame sampling strategies for different temporal transformations.

\textbf{Scale}: The scale transformation in spatial modifies the size of the object along its width or height. Accordingly, we change the duration of an action in temporal scale. To achieve that, we speed up a video by interval sampling, and the interval is saved as the learning target. Specifically, let $n$ be the speed, the interval is $n-1$. That is, given a video $V=\{v_i\}$, where $v_i$ is the $i$-$th$ frame in $V$. We generated $V^s$, which is s times speed of $V$ as $V^{s}=[{v_{r},v_{r+s-1},v_{r+2(s-1)},...,v_{r+l(s-1)}}]$, where $r+l(s-1) \leq$ the length of $V$, $r$ is a random start frame  in $V$ and $l$ is the length of $V^{s}$. In our implementation, $l = 16$. For temporal scale, the speed $s$ is the learning target. Fig. \ref{fig:method} shows an example of $s=2$.

\textbf{Projection}: The spatial projection scales objects in different regions of the video frame at different rates. Accordingly, we propose temporal projection to progressively speed up a video so that it has different playback rates in different stages. In our implementation, we use a multi-stage speed-up mode, which means the video has multiple  speeds in the same video. To generate the training samples, when a video $V$ is given, we first sample $l_1$ frames at $s_1-1$ intervals, then $l_2$ frames at $s_2-1$ intervals. The total length of $V^p$ $l = l_1 + l_2 $. In our implementation, $l_1 = l_2 = 8$. Given a specific speed pattern $p = (s_1,s_2)$ and $V$, we generate $V^p$ as $V^{s_1,s_2}=[{V^{S_1},V^{S_2}}]$. The generation of $V^{S_1},V^{S_2}$ is same as the operation in temporal scale. For temporal projection learning, $p$ is 
 the learning target. Fig. \ref{fig:method} shows an example of $p = (1,2)$.

With temporal transformation, we speed up the video straightforwardly (scale) or progressively (projection), which encourages the model to capture rich temporal representations.

\subsection{Representation Learning}

Given a video clip, we first apply a spatial transformation and then a temporal transformation on it. Then we feed it to a backbone to extract features and use a multi-task network to predict the specific transformation.

\textbf{Feature Extraction:}
To extract video representations, we choose C3D \cite{tran2015learning}, R3D, R(2+1)D \cite{tran2018closer}, and S3D-G \cite{xie2018rethinking} as backbones. C3D stacks five 3D convolution blocks with 3$\times$3$\times$3 convolution kernels in all layers. Within the framework of residual learning, R3D block consists of two 3D convolution layers followed by batch normalization and ReLU layers. R(2+1)D are ResNets with (2+1)D convolutions, which decompose full 3D convolutions into a 2D convolution followed by a 1D convolution. 
Unlike many other 3D CNNs, S3D-G replaces many of the 3D convolutions, especially the 3D convolutions at the bottom of the network, by low-cost 2D convolutions. S3D-G exhibits distinguished feature extraction ability for action recognition.

\textbf{Category Prediction:}
To complete the prediction, we take it as a classification task. For each transformation, we fix the parameters and take them as a specific category for classification. To be noted, to make most use of spatial and temporal transformation, we take V3S as a multi-task network.

Given the feature extracted by 3D CNNs, it is then fed to two fully connected (FC) layers, which completes the prediction. The output of each FC layer is a probability distribution over different categories.  With $a_i$ is the $i$-$th$ output of the fully connected layer for transformation, the probabilities are as follows:
$$
p_i=\frac{\exp(a_i)}{\sum^{n}_{j=1}\exp(a_j)}
$$
\noindent where $p_i$ is the probability that the transformation belongs to class $i$, and $n$ is the number of transformations. We update the parameters of the network by minimizing the regularized cross-entropy loss of the predictions:
$$   
\mathcal{L} = -\sum^n_{i=1}y_i\log(p_i)
$$
\noindent where $y_i$ is the groundtruth.  Let $\mathcal{L^S}$ be the entropy loss for spatial transformation prediction and   $\mathcal{L^T}$ be the loss for temporal transformation prediction. The objective function of V3S is :
$$    \mathcal{L}_{\mathcal{ST}} = \mathcal{L^S} + \mathcal{L^T}
$$

\begin{figure}[!t]     \centering
     \includegraphics[width=0.9\columnwidth]{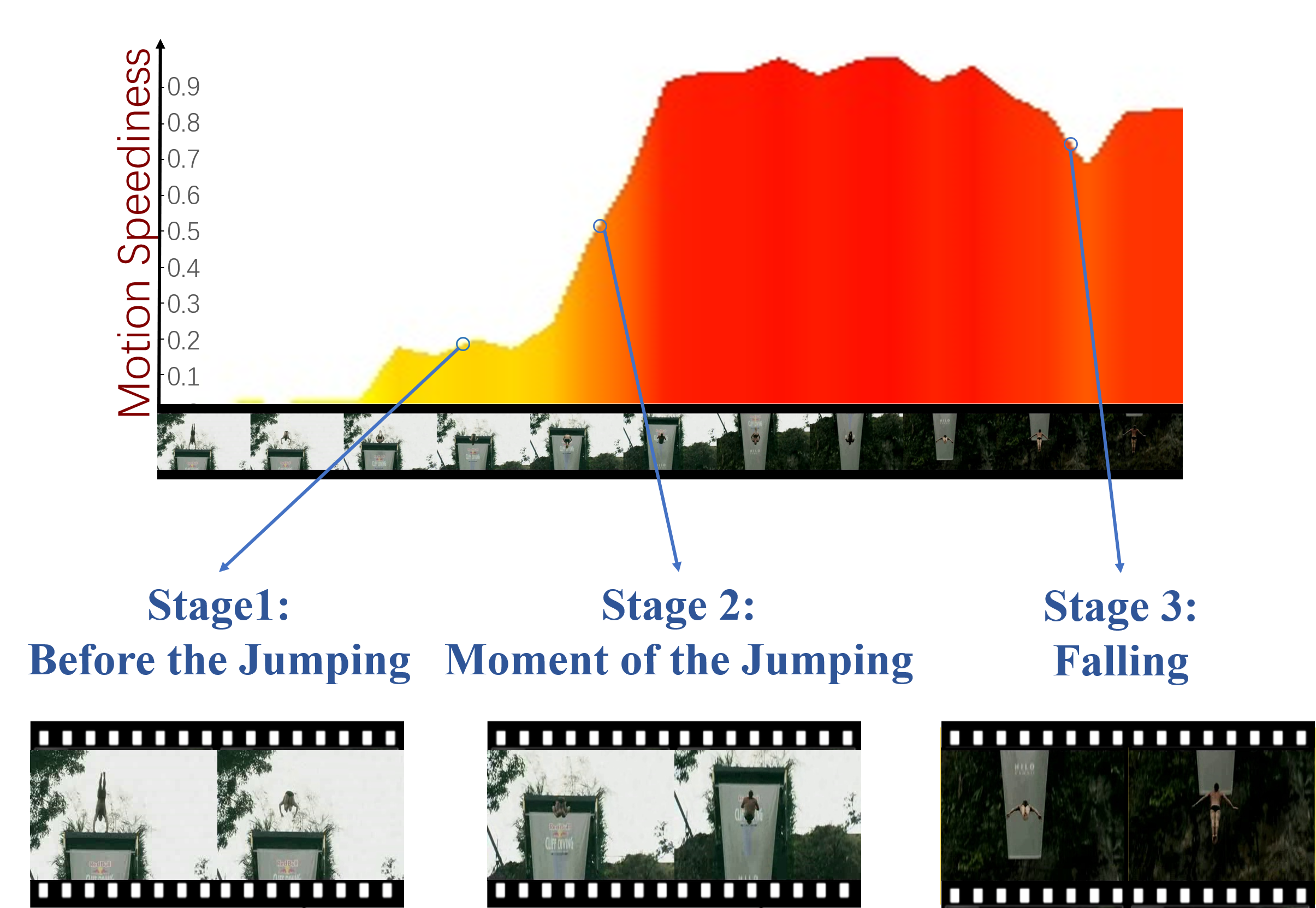}
     \caption{
     Visualization of the motion speediness curve of the cliff diving action video. Cliff diving action can be divided into 3 stages: 1. Before the jumping, the motion speediness curve is constant around 0; 2. Moment of jumping, the speediness curve rises rapidly from 0 to 0.9; 3. Falling, the speediness curve constantly fluctuates around 0.9. Our V3S tries to capture the speed patterns at different action stages.
     }
     \label{fig:speed}
 \end{figure}

\subsection{Discussion}
\label{sec:discussion}
In Fig. \ref{fig:speed}, we show the motion speediness curve \cite{benaim2020speednet} of cliff diving action. As one can see, the speediness of the action is different in different stages, rather than a constant value. Take the cliff diving action as an example, there is a rapid rise of the motion speediness from stage 1 to stage 3. We argue that if the network can sense the variation of speed, it can better understand the characteristics of an action. V3S proposes to utilize temporal projection to capture the variation of speed, thereby completes the deficiencies in  previous speed-based methods.



\section{Experiment}
\label{sec:Experiment}
\subsection{Experimental Setting}
\subsubsection{Datasets}

In our experiments, we use four datasets: the  UCF101 \cite{soomro2012ucf101}, the HMDB51 \cite{kuehne2011hmdb},
the Kinetics-400(K-400)\cite{kay2017kinetics} and the ASLAN \cite{kliper2011action} to evaluate the effectiveness of our method. 

UCF101 is a widely used dataset in  action recognition task, which is collected from websites including Prelinger archive, YouTube and Google videos. The dataset contains 101 action categories with 9.5k videos for training and 3.5k videos for testing. 

HMDB51 is extracted from a variety of sources ranging from digitized movies to YouTube, which contains 3.4k videos for training and 1.4k videos for testing with 51 action categories. 

Kinetics-400 contains 246K train videos. The videos are collected from  realistic YouTube videos. 

ASLAN is an action similarity labeling dataset. It  includes 3,631 videos in over 400 action categories. The goal of action similarity labeling task is to estimate if two videos present the same action or not, and this dataset is composed of video pairs with "same" or "not-same" labels. This task is challenging because its test set only contains videos of  never-before-seen actions. We use this task to verify the spatio-temporal feature extraction capabilities of our model.


\begin{table}
    
    \begin{tabular}{lll}
    \toprule
    Method &Value& Acc.\\
    \midrule
    $S_S$ & \{(1,1.15), (1,1.3), (1,1.45), (1.15,1), (1.3,1), (1.45,1)\} & 73.2\\
    $S_S$ & \{(1,1.3), (1,1.6), (1,1.9), (1.3,1), (1.6,1), (1.9,1)\} & 71.1\\
    $S_S$& \{(1,1.45), (1,1.9), (1,2.35), (1.45,1), (1.9,1), (2.35,1)\} & 72.8\\
        \midrule

    $S_P$ & \{0.8, 0.75, 0.7\} & 73.5\\
    $S_P$ & \{0.8, 0.7, 0.6\} & 73.6\\
    $S_P$ & \{0.8, 0.65, 0.5\} & 73.7\\

    \midrule

    $T_S$& \{1, 2\} & 71.0\\
    $T_S$  & \{1, 2, 3\} & 76.4\\
    $T_S$ & \{1, 2, 3, 4\} & 76.3\\
        \midrule

$T_P$  & \{(1,2) ,(2,3), (2,1), (3,2) \} & 75.0\\
    $T_P$ & \{(1,2), (2,3), (3,4), (2,1), (3,2), (4,3)\} & 76.8\\
    $T_P$ &\{(1,2), (2,3), (3,4), (4,5), (2,1), (3,2),(4,3), (5,4)\} & 77.0\\
   \bottomrule
        \end{tabular}    
        
    \caption{Evaluation of V3S with R(2+1)D under different parameters. $S_S$ denotes spatial scale, $S_P$ spatial projection, $T_S$ temporal scale, $T_P$ temporal projection.}
    \label{table:ablation}
\end{table}

\begin{table*}
    \centering
    \begin{tabular}{lcccccc}
    \toprule
    Method  & Network & Input Size & Params & Pre-train Dataset & UCF101& HMDB51\\
    \midrule
   Random&C3D& 112 × 112&58.3M&UCF101& 63.7 & 24.7\\
   VCP\cite{luo2020video}&C3D&112 × 112&58.3M&UCF101& 68.5 & 32.5 \\
   PRP\cite{yao2020video}&C3D&112 × 112&58.3M&UCF101& 69.1 & 34.5 \\
   V3S(Ours) &C3D&112 × 112&58.3M&UCF101& \textbf{74.8} & \textbf{34.9}\\
      
   \midrule
   Random&R3D&112 × 112&33.6M&UCF101& 54.5 & 23.4\\
  ST-puzzle\cite{kim2019self}&R3D&224 × 224&33.6M&Kinetics-400&65.8 & 33.7 \\
   VCP\cite{luo2020video}&R3D&112 × 112&33.6M&UCF101& 66.0 & 31.5\\
   PRP\cite{yao2020video}& R3D&112 × 112&33.6M&UCF101& 66.5 & 29.7\\
   V3S(Ours)&R3D&112 × 112&33.6M&UCF101 & \textbf{74.0} & \textbf{38.0}\\
  \midrule
   Random&R(2+1)D&112 × 112&14.4M&UCF101& 55.8  & 22.0\\
   VCP\cite{luo2020video}&R(2+1)D&112 × 112&14.4M&UCF101& 66.3 & 32.2\\
   PRP\cite{yao2020video}&R(2+1)D&112 × 112&14.4M&UCF101 & 72.1 & 35.0\\

   PacePred\cite{wang2020self} w/o Ctr&R(2+1)D&112 × 112&14.4M&UCF101& 73.9 & 33.8\\ PacePred\cite{wang2020self}&R(2+1)D&112 × 112&14.4M&UCF101& 75.9 & 35.0\\ 
    PacePred\cite{wang2020self}&R(2+1)D&112 × 112&14.4M&Kinetics-400& 77.1 & 35.0\\
      

   V3S(Ours)&R(2+1)D&112 × 112&14.4M&UCF101 & 79.1  & 38.7\\
   V3S(Ours)&R(2+1)D&112 × 112&14.4M&Kinetics-400 &  \textbf{79.2} & \textbf{40.4}\\
   \midrule
   
   SpeedNet\cite{benaim2020speednet} &S3D-G&112 × 112&9.6M&Kinetics-400&81.1 & 48.8\\
   
    CoCLR\cite{han2020self} &S3D&128 × 128&9.6M&UCF101&81.4 & 52.1\\
    
   
   V3S(Ours) &S3D-G&112 × 112&9.6M&UCF101& \textbf{85.4} & \textbf{53.2} \\

   \bottomrule
   
        \end{tabular}
        
    \caption{Action recognition accuracy on UCF101 and HMDB51.}
    \label{table:action-recognition}
\end{table*}

\subsubsection{Implementation Details} 
For 3D sampling, we firstly transform the raw video and then sample a 16-frame clip from it. Each frame is resized to 224 $\times$ 224 and randomly cropped to 112 $\times$ 112. Specially for spatial projection, if the head end is shorter than 224, we simply crop the frame by $l \times l$, where $l$ is the length of the head end, then it is resized to 112 $\times$ 112.  

We set the initial learning rate to be 0.01, momentum to be 0.9, and batch size to be 32. Our pre-training process stops after 300 epochs and the best validation accuracy model is used for downstream tasks. Specially, to match the requirement of S3D-G, each frame is firstly resized to 256 $\times$ 256, then randomly cropped to 224 $\times$ 224.  In addition, we set the learning rate to be 0.005 for R3D for better convergence. For UCF101, the training set of the first split is used in our pre-training stage, where we randomly choose 800 videos for validation. For Kinetics-400, we use its training set to train  our self-supervised model, and randomly select 3000 samples to build the validation set. The batch size with Kinetics-400 is 16 and we train it for 70 epochs.

    
        

\begin{table}
    
    \begin{tabular}{l|ccc|ccc|c}
    \toprule
    Method  & $S_S$&$S_P$&$S_S$+$S_P$&$T_S$&$T_P$&$T_S$+$T_P$&V3S \\
    \midrule
    Acc. &73.2&73.7&73.8&76.4&77.0&78.0&79.1 \\
   \bottomrule
        \end{tabular}
        
    \caption{Combining spatial and temporal transformations. V3S denotes $S_S+S_P+T_S+T_P$.}
    \label{table:ab2}
    \vspace{-2em}
\end{table}

\subsection{Ablation Study}
In this section, we evaluate the effectiveness and discuss the hyperparameters of the designed four transformations on the first split of UCF101. For simplicity, we choose R(2+1)D as the backbone for our ablation studies. 

As shown in table \ref{table:ablation}, we conduct extensive experiments for the selection of each parameter. In table \ref{table:ablation}, we discuss the influence of hyperparameters for spatial transformations. To explain it further, we discuss $(a,b)$ for spatial scale $S_S$, and $c$ for spatial projection  $S_P$.  For $S_S$ , we randomly select $(a,b)$ $\in$ \{(1,1.15), (1,1.3), (1,1.45), (1.15,1), (1.3,1), (1.45,1)\} in the following experiments, because it demonstrates the best performance among the settings (73.2\% to 71.1\%$\backslash$72.8\%).   For $S_P$, we accordingly select projection magnitude $c$ $\in$ \{0.8, 0.65, 0.5\} in the following experiments. 

In table \ref{table:ablation} (bottom),  we discuss the hyperparameters for temporal transformations, which are $s$ for $T_S$ and $p$ for $T_P$.  For $T_S$ , we randomly select $s$ from $\{1, 2\}$, $\{1,2,3\}$ or $\{1, 2, 3, 4\}$.  When sampling $s$ $\in$ $\{1,2,3\}$, it demonstrates the best performance (76.4\% to 71.0\%$\backslash$76.3\%). we thus set a sampling speed $s \in \{1, 2, 3\}$ in the following experiments. For $T_P$, we accordingly select the speed pattern $p$ $\in$ \{(1,2), (2,3), (3,4), (4,5), (2,1), (3,2), (4,3), (5,4)\} in the following experiments.

To combine the designed transformations, we integrate $S_S$, $S_P$, $T_S$, $T_P$ in Table \ref{table:ab2}. After combining spatial scale $S_S$ and spatial projection $S_P$, V3S achieves 73.8\%. V3S also show better performance (78.0\%) when combining $T_S$ and $T_P$. 
After combining all these four spatial and temporal transformations, the accuracy of 79.1\% is achieved, which surpasses the accuracy of the individual spatial or temporal transformations.
In summary, the spatial and temporal transformations are complementary, thus with the combination as the final proxy task, more powerful representations can be learned.

\begin{table}
    \centering
    \resizebox{\columnwidth}{!}{
    \begin{tabular}{lccccc}
       \toprule
    Method  &Top1&Top5&Top10&Top20&Top50  \\
   \midrule
    Jigsaw & 19.7 & 28.5 & 33.5 & 40.0 & 49.4\\
OPN & 19.9  & 28.7  & 34.0 & 40.6   & 51.6\\
    $\mathrm{B\ddot{u}chler}$&  25.7 & 36.2 & 42.2 & 49.2 & 59.5\\
  \midrule
    C3D(random) & 16.7 & 27.5 & 33.7 & 41.4 & 53.0\\
    C3D(VCP\cite{luo2020video}) & 17.3 & 31.5 & 42.0 & 52.6 & 67.7\\
    
    C3D(PRP\cite{yao2020video}) & \textbf{23.2} & 38.1 & 46.0 & 55.7 & 68.4\\
    C3D(PacePred\cite{wang2020self})& 20.0 & 37.4 & 46.9 & \textbf{58.5} & \textbf{73.1}\\  
    C3D(V3S) & 21.8 & \textbf{39.0} & \textbf{47.7} & 57.0 & 69.2\\
  \midrule
       R3D(random) & 9.9 & 18.9 & 26.0 & 35.5 & 51.9\\
    R3D(VCP\cite{luo2020video}) & 18.6 & 33.6 & 42.5 & 53.5 & 68.1\\
    R3D(PRP\cite{yao2020video}) & 22.8 & 38.5 & 46.7 & 55.2& 69.1\\
    R3D(PacePred\cite{wang2020self}) & 19.9&36.2&46.1&55.6&69.2\\
    R3D(V3S) & \textbf{28.3} &  \textbf{43.7} & \textbf{51.3} & \textbf{60.1} & \textbf{71.9}\\
  \midrule
    R(2+1)D(random) &  10.6 & 20.7 & 27.4 & 37.4 & 53.1\\
   R(2+1)D(VCP\cite{luo2020video}) & 19.9 & 33.7 & 42.0 & 50.5 & 64.4\\
   R(2+1)D(PRP\cite{yao2020video}) & 20.3 & 34.0 & 41.9 & 51.7 & 64.2\\
    R(2+1)D(PacePred\cite{wang2020self})&17.9&34.3&44.6&55.5&72.9\\
   
    R(2+1)D(V3S) & 23.1 & \textbf{40.5} &48.7 & 58.5 &72.4\\
    R(2+1)D(V3S*) & \textbf{23.5} & 40.0 &\textbf{49.4} & \textbf{59.7} & \textbf{73.9}\\
    
  \midrule
    S3D-G(SpeedNet*\cite{benaim2020speednet}) & 13.0& 28.1 & 37.5 &49.5 &
    65.0\\
    
    S3D-G(V3S) & \textbf{16.6} & \textbf{32.2} & \textbf{41.8} & \textbf{52.3} &\textbf{68.0}\\

    \bottomrule
        \end{tabular}
        }
    \caption{Video retrieval performance on UCF101. Methods   marked   with   *   are   pretrained with Kinetics-400.}
    \label{table:retrieval-UCF}
    \vspace{-1cm}
\end{table}

\subsection{Action Recognition}
Utilizing self-supervised pre-training to initialize action recognition models is an established and effective way for evaluating the representation learned via self-supervised tasks. To verify the effectiveness of our method, we conduct experiments on the action recognition task. We initialize the backbone with V3S pre-trained model, and initialize the fully connected layer randomly. Following the protocol of \cite{xu2019self}, we train backbones for 300 epochs during training and make the fine-tuning procedure stop after 160 epochs. For testing, we sample 10 clips for each video and average the possibility of predictions to obtain the final action category.

For C3D, R3D and R(2+1)D, we set the initial learning rate to be 0.001. The number of the clip frames is 16  and each frame is first resized to 128 $\times$ 171 and randomly cropped to 112 $\times$ 112.  For S3D-G, we set the initial learning rate to be 0.01, the input clip length is 64 frames and each frame is first resized to 256 $\times$ 256 and randomly cropped to 224 $\times$ 224. For action recognition, the batch size is set to 8 and the momentum is set to 0.9.

Table \ref{table:action-recognition} shows the split-1 accuracy on UCF101 and HMDB51 for action recognition task. With S3D-G pretrained on UCF101, V3S obtains 85.4\% and 53.2\% on UCF101 and HMDB51 respectively, outperforms CoCLR \cite{han2020self} by 4.0\% and 1.1\%.  
With R(2+1)D pretrained on Kinetics-400, V3S achieves 79.2\% and 40.4\% on UCF101 and HMDB51, which outperforms PacePred\cite{wang2020self} by 2.1\% and 5.4\%.
With C3D, R3D, V3S obtains  74.8\%$\backslash$34.9\% and 74.0\%$\backslash$38.0\%. 
To be noted, V3S is purely pretext task based, and does not incorporate contrastive learning. However, its performance is better than a series of methods with contrastive loss  such as PacePrediction\cite{wang2020self} and CoCLR\cite{han2020self}. This can further verify its effectiveness. 

\begin{table}

    \centering
    \resizebox{\columnwidth}{!}{
    \begin{tabular}{lccccc}
       \toprule
    Method  &Top1&Top5&Top10&Top20&Top50\\
    \midrule
    C3D(random) & 7.4 & 20.5  & 31.9 & 44.5 & 66.3\\
    C3D(VCP\cite{luo2020video}) & 7.8 & 23.8 & 35.5& 49.3 & 71.6\\
    C3D(PRP\cite{yao2020video}) & \underline{10.5} & \textbf{27.2 }&\textbf{40.4} & \textbf{56.2} & \underline{75.9}\\
    C3D(PacePred\cite{wang2020self}) &8.0&25.2&37.8&\underline{54.4}&\textbf{77.5}\\

    C3D(V3S) & \textbf{11.1} & \underline{26.5} & \underline{38.0} & 52.0 & 73.0\\
    \midrule
    R3D(random) & 6.7 & 18.3 & 28.3 & 43.1 & 67.9\\
    R3D(VCP\cite{luo2020video}) & 7.6 & 24.4 & 36.6 & \underline{53.6} & \underline{76.4} \\
    R3D(PRP\cite{yao2020video}) & \underline{8.2} & \underline{25.8} & \underline{38.5} & 53.3 & 75.9\\
    R3D(PacePred\cite{wang2020self}) &\underline{8.2}&24.2&37.3&53.3&74.5\\
    R3D(V3S) &\textbf{10.8} & \textbf{30.6} & \textbf{42.3 }& \textbf{56.2} & \textbf{77.1}\\
    \midrule
    R(2+1)D(random) &  4.5 & 14.8 & 23.4 & 38.9 & 63.0\\
   R(2+1)D(VCP\cite{luo2020video}) & 6.7 & 21.3 & 32.7 & 49.2 & 73.3\\
   R(2+1)D(PRP\cite{yao2020video}) & 8.2 & \underline{25.3} & 36.2 & 51.0 & 73.0\\
   
   R(2+1)D(PacePred\cite{wang2020self}) &\textbf{10.0}&24.6&\underline{37.6}&\textbf{54.4}& \underline{77.1}\\
   
    R(2+1)D(V3S) & 9.6 & 24.0 & 37.2 & \underline{54.3} &\textbf{77.9}\\
    R(2+1)D(V3S*) & \underline{9.8} & \textbf{26.9} & \textbf{38.5} & 52.7 & 72.2\\
    \bottomrule

        \end{tabular}
        }
    \caption{Video retrieval performance on HMDB51. Methods   marked   with   *   are   pretrained   with   Kinetics-400. The best results are bold, and the second best results are underlined.}
    \label{table:retrieval-HMDB}
\end{table}

\subsection{Video Retrieval}
To further validate the effectiveness of V3S, we adopt video retrieval as another downstream task. In the process of retrieval, we generate features at the last pooling layer of extractors. For each clip in the testing split, we query top-K nearest videos from the training set by computing the cosine similarity between two feature vectors. When the testing video and a retrieved video are from the same category, a correct retrieval is counted.

Video retrieval results on UCF101 and HMDB51 are listed in Table \ref{table:retrieval-UCF} and Table \ref{table:retrieval-HMDB} respectively. Note that we outperform SOTA methods with different backbones from Top1 to Top50. These results indicate that in addition to providing a good weight initialization for the downstream model, V3S can also extract high-quality and discriminative spatiao-temporal features.  
\begin{table}
     \begin{tabular}{lcccc}
    \toprule
    Features  & Hand-crafted & Sup.&Self-sup.&Acc.\\
    \midrule
    C3D&&\checkmark&&78.3\\
    \midrule
    HOF&\checkmark&&&56.7\\
    HNF&\checkmark&&&59.5\\
    HOG&\checkmark&&&59.8\\
    \midrule
    STS\cite{wang2019self},R3D&&&\checkmark&60.9\\
    V3S,R3D&&&\checkmark&\textbf{65.4}\\
    
   \bottomrule
        \end{tabular}    
    \caption{Action similarity accuracy on ASLAN.}
    \label{table:Action similarity}
\end{table}

\subsection{Action Similarity Labeling}
In this section, we exploit action similarity labeling  task \cite{kliper2011action}  to verify the quality of the learned spatio-temporal representations from another perspective on the ASLAN dataset\cite{kliper2011action}. Unlike action recognition task, the action similarity labeling task focuses on action similarity (same/not-same). The model needs to determine whether the action categories of the two videos are the same. This task is quite challenging as the test set contains never-before-seen actions.

\begin{figure*}
     \centering
     \includegraphics[width=2.0\columnwidth]{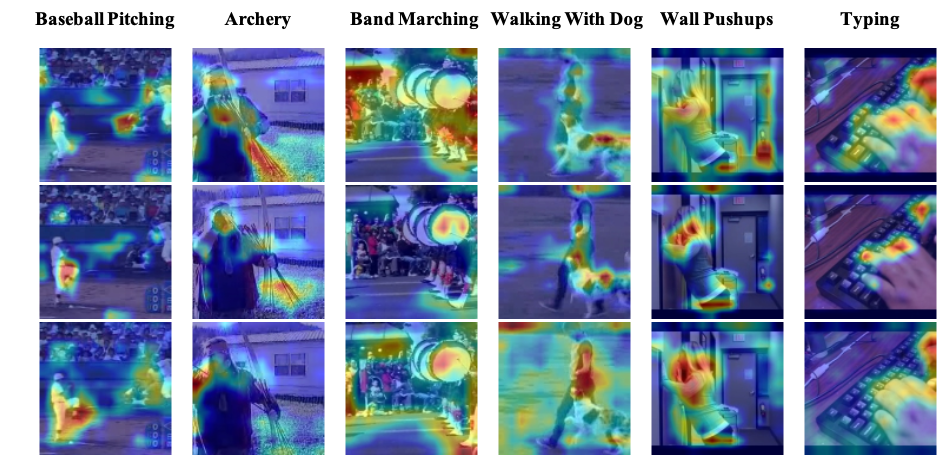}
     \caption{Attention maps of video frames.}
     \label{fig:attention}
\end{figure*}

To evaluate on the action similarity labeling task, we use pre-trained model to extract features from video pairs, and use a linear SVM for the binary classification.  Specifically, for each video pairs, each video is first split into 16-frame clips with stride length 8.
The network takes these video clips as input to extract clip-level features from the pool3, pool4 and pool5 layers. Then, the clip-level features are averaged and L2 Normalized to get a spatiao-temporal video-level feature.
In order to measure the similarity of these two video-level features, we calculate  12 different distances for each feature as described in \cite{kliper2011action}. A 36-dimensional feature is obtained by concatenating these  three types of video features and this feature is normalized to ensure that the scale of each distance is the same, following \cite{tran2015learning}. Finally, a linear SVM is used to determine whether the two videos are in the same category or not.

Action similarity results on ASLAN are listed in Table \ref{table:Action similarity}. The results show that the accuracy of V3S on R3D outperforms that of the previous methods, and it further shortens the gap between supervised and unsupervised methods. 
It demonstrates that the features extracted by the V3S network have excellent intra-class similarity and inter-class dissimilarity.




\subsection{Visualization}
In order to gain a better understanding of what V3S learns, we adopt the Gradient-weighted Class Activation Mapping (Grad-CAM) {\cite{selvaraju2017grad}}, an improved version of CAM\cite{zhou2016learning} to visualize the attention map. 

Fig. \ref{fig:attention} shows the samples (baseball pitching, archery, band marching, baseball pitch, walking with dog, wall pushups, typing) of such heat maps. One can see that the highly activated regions of these heat maps have a great correlation with the movement of actions. For example, in different stages of the archery action, the activation area varies along the motion. When the man takes out of the arrow, the activation area concentrates on the left hand that is drawing the arrow, and when the man is drawing the bow, the activation area moves to the right hand.

\section{Conclusion}
In this paper, we propose a novel self-supervised method referred to as V3S to obtain rich spatio-temporal features without human annotations. In V3S, to fully utilize the information in videos, we propose spatial scale and spatial projection to uniformly or non-uniformly scale the objects in a video. We propose temporal scale and temporal projection to straightforwardly or progressively speed up a video.  Experimental results show the effectiveness of V3S for downstream tasks such as action recognition, video retrieval and action similarity labeling. Our work inspires the field of video understanding with two aspects: non-contrastive learning based self-supervised pretext task learning is still below the upper bound, and powerful representations can be learned with relatively small datasets like UCF101.








\bibliographystyle{ACM-Reference-Format}
\bibliography{sample-base}

\appendix

\end{document}